# Toward using explainable data-driven surrogate models for treating performance-based seismic design as an inverse engineering problem

## Mohsen Zaker Esteghamati[*]

*Department of Civil & Environmental Engineering, Utah State University, 4110 Old Main Hill, Logan, UT 84321, USA, ORCID: 0000-0002-2144-2938*

***Keywords:*** *Machine learning, surrogate modeling, performance-based design, inverse problems, explainability*

## Summary

This study presents a methodology to treat performance-based seismic design as an inverse engineering problem, where design parameters are directly derived to achieve specific performance objectives. By implementing explainable machine learning models, this methodology directly maps design variables and performance metrics, thereby tackling the computational inefficiencies associated with performance-based design. The resultant machine learning model is integrated as an evaluation function into a genetic optimization algorithm to solve the inverse problem. The developed methodology is then applied to two different inventories of steel and concrete moment frames in Los Angeles and Charleston to obtain sectional properties of frame members that minimize expected annualized seismic loss in terms of repair costs. The results show high accuracy of the surrogate models (e.g., $R^2 > 90\%$) across a diverse set of building types, geometries, seismic design, and site hazard, where the optimization algorithm could identify the optimum values of members' properties for a fixed set of geometric variables, consistent with engineering principles.

## 1. Introduction

Performance-based seismic design (PBSD) characterizes a structural system (e.g., member proportioning, selection of member sections) to achieve specific performance objectives (PO), where each PO represents the coupling of performance level (e.g., damage or consequence-related metrics) and seismic hazard level [1]. A reliable PBSD requires defining multiple POs, where the definition and number of selected POs depend on the project needs, regulations, and involved stakeholders. Despite the potential of PBSD for infrastructure resiliency-based applications, significant challenges exist in its formal implementation as part of a routine design procedure, which have hindered its broader applications. These challenges include algorithmic issues, data unavailability, the effort- and time-intensive nature of the required analyses,

*[*]Mohsen Zaker Esteghamati (mohsen.zaker@usu.edu).*

*[†]Present address: Department of Civil & Environmental Engineering, Utah State University, 4110 Old Main Hill, Logan, UT 84321, USA*

and the accrual of different sources of uncertainties that affect design outcomes [2], collectively motivating the development of novel computational paradigms and methods to streamline this process.

PBSD, similar to other structural design problems, is mainly treated as a forward problem that determines system output (performance; $y$) for a set of system inputs (i.e., design solutions; $x$), where the designer iteratively changes the input to converge to the desired output (i.e., $x \rightarrow y$). In contrast, an inverse problem formulation [3,4] identifies system characteristics that result in specific system outputs by establishing a mathematical relationship between input and output (i.e., $y \rightarrow x$). Unlike forward models, inverse problems are often ill-posed and may have no or multiple solutions, or cannot be directly calculated by inverting a forward problem due to errors in response measurements. Therefore, such an inverse problem perspective is closely aligned with PBSD objectives, which conform to the shared qualities of inverse problems, such as the non-uniqueness of the solution, particularly for more complex systems or across multiple objectives [5]. Additionally, formulating PBSD as an inverse problem facilitates efficient algorithmic implementation due to the extensive literature on inverse problems and the availability of various computational methods for their solution. Common approaches to solving inverse problems include the direct calculation of $x$ by creating a mapping ($f^1$) of $y \rightarrow x$ (i.e., $f^1$ ($y$)= $x$) through analytical or data-driven surrogates [6,7], gradient-based iterative methods [8,9], Bayesian methods [10,11], and optimization-based approaches [12,13].

Machine learning (ML) offers a computationally efficient solution for constructing an inverse problem and exploring the solution space [14–17]. To this end, ML models are leveraged to provide direct mapping ($f$), solve part of the optimization problem (e.g., providing a surrogate to calculate the forward problem), or replace the entire optimization problem [18]. Treating PBSD as an inverse problem provides a basis to guide ML model development, and direct integration of multiple design objectives and structural optimization literature. Nevertheless, most ML-based studies have focused on estimating response or performance (i.e., the forward problem) using detailed inputs that are often unavailable during a typical design task. Among the few studies that primarily investigate ML as a design solution, Gallet *et al.* applied a neural network to predict the cross-sectional properties of continuous beam systems with arbitrary sizes through a non-iterative approach based on the influence zone concept. Gu *et al.* focused on using diffusion models to optimize shear wall layouts and improve building drift responses [19]. Further work in this domain has focused on improving model understanding of component relationships and their performance using sparse features [20]. Zaker Esteghamati and Flint [21,22] (and later, Zaker Esteghamati and Baddipalli [23]) investigated ML-based surrogate models to construct a mapping between design and seismic performance endpoints, demonstrating the relative capabilities of these approaches.

This study presents an ML-based framework to handle PBSD as an inverse problem by combining optimization techniques with data-driven surrogate models trained on performance-based simulation data. The underlying ML models are referred to as "design-centric" models, distinguishing them from the common ML models that solely focus on constructing forward problems. Design-centric ML models have three key characteristics: (1) provide a direct mapping between performance endpoints and design parameters of a structural system, (2) careful treatment of model input to ensure the availability of input features during the design process, and compatibility with design workflows, and (3) high interpretability (or the ability to be explained) to improve the confidence in the constructed mapping. This study is structured as follows: Section 2 provides an overview of the presented methodology; Section 3 describes the illustrative examples of two databases of steel and concrete frames, including data, methodology, and algorithms, and lastly, Section 4 presents the results and discussion of the applied methodology.

## 2. Methodology

### 2.1. Overview

The key component of this methodology is to use an ML-based surrogate model to estimate $f(x)$ based on training data (i.e., $\hat{f}(x)$) and uses an optimization algorithm to obtain likely $x$ (i.e., design parameters for a given building type and geometry at a fixed location) that will lead to specific values of $y$ (e.g., performance metrics, such as cost or time of recovery) as follows:

$$x^* = \arg\min_x \left\{ L\left(y, \hat{f}(x)\right) + \lambda R(x) \right\} \tag{1}$$

where $L$ is the loss function, $\lambda R(x)$ denotes a regularization term to address potential ill-posedness and instability issues, and to mitigate the risk of overfitting. Figure 1 shows the schematics of the proposed methodology. Here, a genetic algorithm (GA) was used to solve this inverse problem, although other computational approaches can be used to solve the inverse problem. GA is an evolutionary metaheuristic algorithm that has been widely applied for complex optimization problems (e.g., large search space) [24,25]. The GA is initialized with a random sample of possible solutions (i.e., individuals), where each individual is then assessed using a fitness evaluation (a measure of the quality of solutions), and the best individuals are selected through fitness-proportionate methods. Next, crossover or mutation operations are performed on the best individuals to yield the next generation of individuals (solutions) through partial or complete replacements. Here, crossover combines the information of the current solutions with the best solutions, whereas mutation introduces random information. Therefore, the efficiency of GA stems from constructing better solutions by combining the best solutions at each generation.

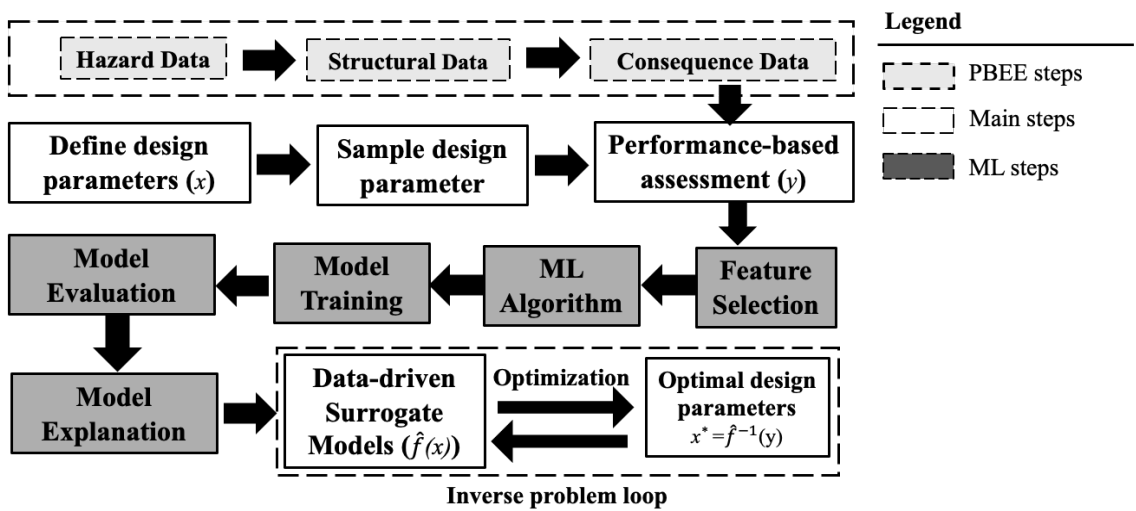

**Figure 1.** Overview of the methodology

As shown in Figure 1, developing suitable ML surrogate models is the most critical task, and requires sampling the design space, creating a performance inventory with adequate fidelity to provide the performance metrics of interest (either point estimate or distribution) over the sampled design scenarios, and train supervised models with minimum input that conforms the level of information available at the time of the design. These ML models, coined as *design-centric* in this study, provide a direct mapping between design and performance, which paves the path toward their direct implementation in an optimization problem. The performance inventory can be compiled by performing detailed performance-based assessments, leveraging open data and simplified tools [26], or from prior knowledge (e.g., databases of published performance data [27,28]). Due to current data limitations, a combination of these approaches should be employed to provide adequate training data for the ML algorithms for most real design scenarios [22]. Lastly, it should be noted that the presented framework uses a gradient-free optimization approach, where ML models serve as a computationally efficient surrogate to calculate the objective function rather than providing gradients for the optimization solver, offering more flexibility due to the high nonlinearity of the design space problem.

## 2.2. Explainability

Explainability has generated a significant literature despite its recency in ML research. Nevertheless, no universal definition or taxonomy exists to encompass all conceptual nuances and methodological advancements of this domain [29]. A key concept for explainability is "model transparency" [30] to support user's improved understanding, and consequently trust, of the model output. Several authors associate "explainability" with "interpretability", whereas some distinguish the two concepts [31–33]. A common definition is that while interpretability focuses on models that are inherently understandable (model's internal mechanism), explainability focuses on "post-hoc" explanations of models' output to users [34]. In the proposed framework, explainability is considered as a post-hoc method to explain how a model derives its prediction, in contrast to ante-hoc methods that focus on simpler models with high algorithmic transparency. The inclusion of the "explainability" requirement provides two key advantages. First, it allows to examine whether the developed mapping function (i.e., design parameter-performance metric) captures true or intuitive engineering principles. Second, these post-hoc explanations facilitate isolating the most important features (e.g. design variables), which could improve computational efficiency for optimization applications and design space exploration.

Two different techniques of Shapley additive models and accumulated local effects (ALEs) were used for explainability in this study. Shapley explainers use cooperative game theory to identify each feature's contribution to the final prediction by calculating Its average marginal contribution across all possible subsets [35]. In contrast, ALEs calculate the variation of model prediction by changing the features Interval and maintaining all other features constant [36]. In this study, Shapley plots are used to understand the most Important features for model prediction, whereas ALE plots are used to construct the relationship between critical features and EALs.

## 2.3. Performance-based seismic assessment

There are various methods to estimate fragility and loss in a PBSD [37]. Here, fragilities were defined using a linear regression-based formula to accommodate data from a cloud analysis as follows:

$$F\left(DS_{aj}\big|EDP\right) = \Phi\left(\frac{\ln(EDP) - \ln\left(\mu_{EDP|DS_{aj}}\right)}{\beta_{EDP|DS_{aj}}}\right) \tag{2}$$

where $\mu_{EDP|DS_{aj}}$ and $\beta_{EDP|DS_{aj}}$ are the median and dispersion EDP demand of $a^{th}$ assembly correspond to $j^{th}$ damage state. The EDP thresholds in Equation.(2) were adopted from HAZUS-MH 2.1 [38]. The expected loss, $E[L_T|NC \cap R, IM]$, was then evaluated as follows:

$$E[L_T|NC \cap R, IM] = \sum_{i=1}^{a}\sum_{j=1}^{n}\int_{0}^{\infty} E\left(L_{aj}\big|DS_{aj}\right)P\left(DS_{aj}\big|EDP\right)f(EDP|IM)dEDP \tag{3}$$

where $L_T$ is the total loss and NC $\cap$ R shows the case where the building has not collapsed but requires repair due to damage. $a$ and $n$ are the number of assemblies considered in the building and the number of damage states of each assembly, respectively. $E\left(L_{aj}\big|DS_{aj}\right)$ shows the expected repair cost of $a^{th}$ assembly at the $j^{th}$ damage state, which was taken from HAZUS-MH 2.1 for steel and RC frames based on their height and occupancy category in terms of building total cost. Building repair costs were then translated into monetary values based on the construction cost databases. The loss values were then converted into expected annual loss (EAL) by multiplying conditional values of loss with IM's exceedance rate, and integrating over possible IM values as follows:

$$EAL = \int E[L_T|IM]\left|\frac{d\lambda_{IM}}{dIM}\right|dIM \tag{4}$$

where $E[L_T|IM]$ is derived from Equation (3) and $d\lambda_{IM}$ represents seismic hazard curve at the site. The EAL values were used as the desired output of ML-based based surrogate models in the next section.

## 2.4 Surrogate mode development

The choice of ML surrogate models depends on the application. Past research showed the potential of support vector machines (SVMs) to create a mapping between design parameter and loss [21,23]. An SVM determines the hyperplane that separates training data points and deviates at a maximum allowable margin ($\varepsilon$). A hyperplane is a flat affine subspace where any data point can be located on either side of this plane. SVMs use soft margins, allowing observation to violate the classifier and, as a result, tackle the possible sensitivities to individual observations that might not be separable by a hyperplane [21,39]. The resulting loss function can be written as follows:

$$J(\Theta) = \frac{1}{2}|w|^2 \text{ subject to } \begin{cases} y_i - \displaystyle\sum_{m=1}^{M} w_m \varphi_m(x) - b \leq \varepsilon \\ \displaystyle\sum_{m=1}^{M} w_m \varphi_m(x) + b - y_i \leq \varepsilon \end{cases} \tag{5}$$

where M is the number of mapping functions of m, $\varphi m$ represents the feature mapping into a higher-dimensional space, and b is the bias term. Equation (5) constraints limit the deviation from target values to $\varepsilon$. Different statistical metrics exist to measure the performance of a data-driven surrogate model. In this study, three common metrics of adjusted $R^2$, normalized RMSE, and normalized MAE were considered as follows:

*Phil. Trans. R. Soc. A.*

$$RMSE_{normalized} = \frac{\sqrt{\frac{1}{n}\sum_{i=1}^{n}(y_i - \widehat{y_i})^2}}{\bar{y}}$$

$$MAE_{normalized} = \frac{\frac{1}{n}\sum_{i=1}^{n}|y_i - \widehat{y_i}|}{\bar{y}} \qquad (6)$$

$$R_{adjusted}^2 = 1 - \left(1 - \frac{\sum_{i=1}^{n}(y_i - \widehat{y_i})^2}{\sum_{i=1}^{n}(y_i - \bar{y})^2}\right) \cdot \frac{n-1}{n-p-1}$$

where $y_i$, $\widehat{y_i}$, and $\bar{y}$ represent actual, predicted, and average value of responses, and $p$ and $n$ show the number of features and data instances. RMSE and MAE are measures of model error, where MAE is robust to outliers and RMSE is sensitive to the spread of data. In contrast, the adjusted R-squared measure indicates the goodness of fit, showing how well a model explains the variance in the response.

## 3. Illustrative case study

This section discusses a case study of framed steel and concrete buildings to illustrate the proposed methodology and its applications.

### 3.1. Performance Inventory Compilation

This study used two different databases of steel and reinforced concrete (RC) frame buildings. The Steel database [40] consists of 621 special moment-resisting designed for Los Angeles, CA, whereas the RC inventory [21] provides 720 special moment-resisting frames for Charleston, SC. The steel frame buildings have five bays in each direction (1 to 5 lateral-resisting bays) with varying spans from 20 $ft$ to 40 $ft$. The number of stories also varied from 1 to 19, with a typical story height of 13 $ft$ and first story height values of 13, 19.5, and 26 $ft$. On the other hand, The RC frame buildings include 2 to 6 bays in each direction with span lengths from 21 $ft$ to 30 $ft$. The number of stories varies between 3 to 6 stories. The typical story height was 12 $ft$, whereas the first story height was considered as 14 $ft$. The RC and steel buildings are exposed to high seismic hazards with $S_{DS}$ = 0.75 $g$ and site class B/C (rock/dense soil) for Charleston, and $S_1$ = 0.6 $g$ and $S_s$ = 2.25 $g$ and site class D for Los Angeles, respectively. Both buildings consisted of regular planar buildings, where the lateral-resisting frames were located around the perimeter of the buildings.

The steel and concrete databases provide floor-level structural responses, ground motion records, and two-dimensional nonlinear finite element models. The nonlinear models were developed in the OpenSees finite element code [41] using a concentrated plastic hinge model. In this approach, the frame members are modeled as elastic members with two plastic hinges at the ends. The properties of the plastic hinges were obtained through regression equations provided by Haselton et al. [42] and Lingos et al. for RC and Steel frames, respectively. A *Joint2D* model was used for the RC frame to simulate concrete connections, whereas a scissor model was applied for steel frame panel zones. In both databases, the leaning column concept was used to apply P-delta effects due to gravity frames. Additional information for frame modeling can be found in [21] and [40], respectively. The RC database provides maximum inter-story and peak floor acceleration under a suite of 80 site-specific ground motions obtained through a geologically realistic probabilistic seismic hazard analysis of Charleston [43] as input for stochastic GM simulation [21]. In contrast, the steel database responses were obtained through 240 unscaled GM records from 12 past events in California, with magnitudes between 6 and 7 and peak ground acceleration values ranging from 0.03 $g$ to 0.61 $g$ [44].

To generate the data for training and testing the surrogate model, a PBEE assessment was performed on both databases, where building seismic loss was calculated in terms of the expected annual repair cost. To this end, an assembly-based approach [45,46] was used where building component loss was aggregated into several assemblies, and the consequence functions of each assembly were related to the building-level fragility functions. An occupancy type of "commercial" was selected for both inventories. These assemblies include structural, drift- and acceleration-sensitive non-structural components. Fragility functions were constructed using a linear regression model in logarithmic space between engineering demand parameters (EDPs) and ground motion intensity measures (IMs) based on the results of cloud analysis for each database. EDP and IM were taken as maximum inter-story and floor acceleration over all stories and spectral acceleration at the building's fundamental period, respectively [47].

## 3.2. Surrogate model development

As shown in Figure 1, the model development workflow comprises feature selection, model training, and model evaluation modules. A two-step feature selection approach [48] was used, where a recursive feature elimination algorithm identified the best $k$-features based on model error for the selected algorithm, and a second step determined the optimum value of $k$ as a trade-off between the number of features and model accuracy. In model training, the algorithm's hyperparameters were tuned using a hybrid search algorithm based on randomized and Bayesian search algorithms, where the prior Bayesian algorithm was constructed using the randomized search algorithm. A 3-fold cross-validation was performed to ensure hyperparameters were not identified for a specific training data sample. In this approach, the training data was partitioned into three subsamples, and the hyperparameters were obtained for two of the subsamples and then tested on the third subsample. This process was repeated three times to yield the best hyperparameters. The tuned SVM model was then used to predict EAL for both data inventories.

The model features comprised geometry- and design-related parameters of steel and RC frames to facilitate their application for the inverse problem of PBSD. For both steel and RC frame databases, the geometry-related features include the number of stories ($NS$), floor area ($A_F$), and bay width ($BW$), whereas the steel database also includes the height of the first story ($h_1$), as these values vary between different buildings. In contrast, the plan aspect ratio ($B/L$) was included in the ML model for RC buildings, as the training database included buildings with unequal dimensions. The design-related parameters were selected based on the system type. For steel frames, these parameters included the average, maximum, and minimum moment of inertia for beams (i.e., $I_{b,avg}$, $I_{b,max}$, and $I_{b,min}$, respectively) and internal (.e., $I_{c,i,avg}$, $I_{c,i,max}$, and $I_{c,i,min}$, respectively) and external (.e., $I_{c,ext,avg}$, $I_{c,ext,max}$, and $I_{c,ext,min}$, respectively) columns. All these statistics were computed over the entire building. In contrast, for RC frames, the design parameters consist of average cross-section area of beams and columns over the entire building ($A_{b,avg}$ and $A_{c,avg}$) and at the first floor($A_{b,1}$ and $A_{c,1}$), and their associated longitudinal reinforcement ratios ($\rho_{b,avg}$, $\rho_{c,avg}$, $\rho_{b,1}$ and $\rho_{c,1}$, respectively). Lastly, total building weight ($W_T$) was also selected as an important seismic-related parameter that is estimable by a designer based on preliminary layouts. The training time was about 26-35 seconds on a Mac M3 Pro Laptop with 12 cores and 36 GB of RAM.

*Phil. Trans. R. Soc. A.*

## 3.3. Genetic algorithm optimization

This study formulates GA to minimize EAL as a proxy of operational life cycle costs associated with earthquake hazard. Each individual (design solution) is defined through 10 and 7 design-related attributes (genes) for steel and RC frames, mimicking a real-world design scenario where the designer knows the geometry, location, and type of the lateral system, and aims to find appropriate design solutions over bounded design input value (minimum and maximum values determined using the databases). A custom gene generator was used to randomly sample genes (design values) within the possible range of design values incorporated into the ML training. A blend crossover with an alpha value of 0.5 was employed, while a Gaussian mutation was applied to each gene, with a 0.2 probability of mutation. Additionally, an adaptive mutation scheme was implemented to adjust the mutation probability over generations, enabling the optimization to focus on more promising solutions in later generations. 10% of top-performing individuals were carried forward to the next generation without mutation or crossover. Lastly, an early stopping was included to terminate the algorithm if no improvements were observed over five consecutive generations. The optimization was performed in about 25 seconds on the system above.

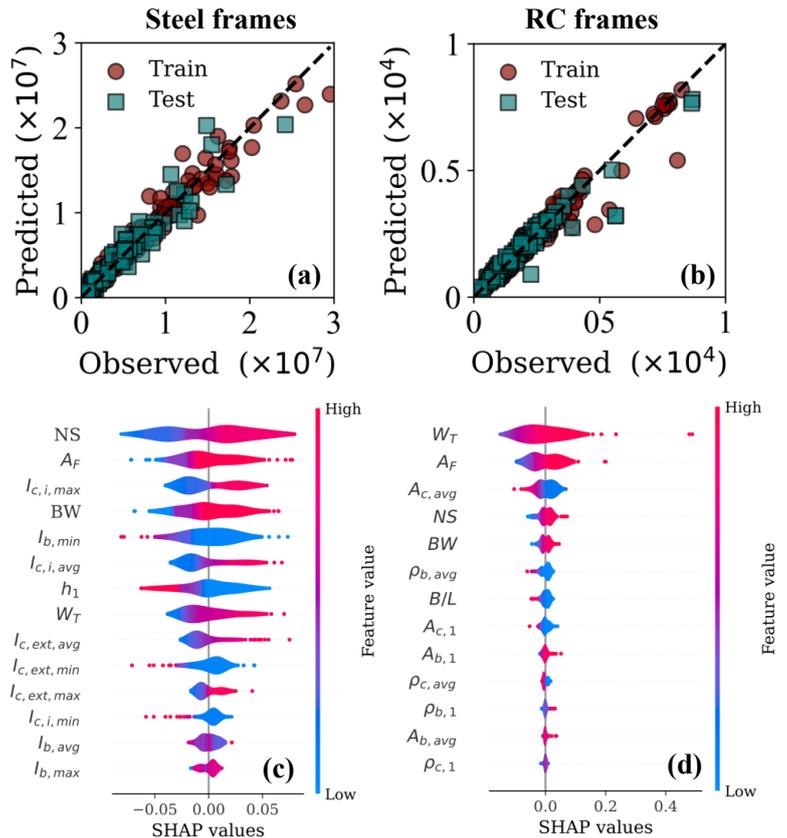

## 4. Results

Figures 2.(a) and 2.(b) compare model performance for training and testing sets of steel and RC building inventories, respectively. Additionally, Table 1 shows the accuracy metrics of the developed surrogate models. Overall, both models show high accuracy, with adjusted $R^2$ values of 92% and 90% on the test set for steel and RC frame building, respectively. The normalized RMSE and MAE values for the steel frame are 0.29 and 0.18, whereas the same metrics are 0.20 and 0.09 for the RC database. The slightly

**Figure 2.** Comparison of developed SVM for (a) steel, and (b) RC frame inventories. Figure (c) and (d) shows the corresponding explanations for each model through Shapley explainer (Shap plots)

lower performance of the steel buildings database can be attributed to its broader scope, which encompasses short, medium, and tall buildings (as opposed to only medium-rise buildings for RC frames) under higher seismicity, and subsequently introduces larger variability in response. Comparing results to the limited published literature on direct mapping of loss to building parameters demonstrates a similar level of accuracy [21,23,49].

**Table 1.** Comparison of SVM surrogate models' accuracy for steel and RC frame building

| Building inventory | Adjusted R² | | Normalized RMSE | | Normalized MAE | |
|---|---|---|---|---|---|---|
| | Training | Testing | Training | Testing | Training | Testing |
| **RC frame** | 96.66% | 90.37% | 0.12 | 0.20 | 0.06 | 0.09 |
| **Steel frame** | 97.28% | 92.09% | 0.19 | 0.29 | 0.10 | 0.18 |

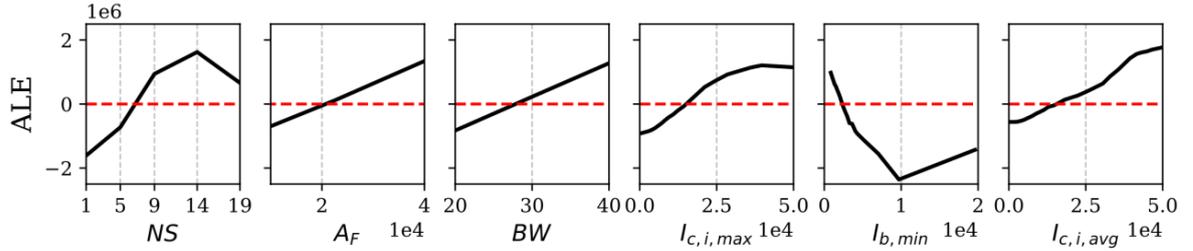

(a) Steel frames

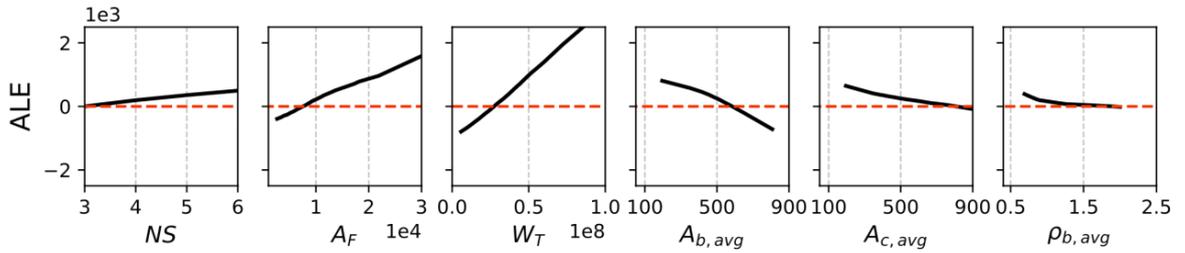

(a) RC frames

**Figure 3.** Accumulated local effect (ALE) plots for (a) steel and (b) RC frames for influential parameters

Figures 2.(c) and 2.(d) show the Shapley explanations for EAL of steel and RC frames, respectively. Overall, the same geometric parameters of $NS$, $A_F$, and $BW$ have the largest impact on EAL of both RC and steel frames. In addition, $I_{c,i,max}$, and $I_{b,min}$ are among the most influential parameters affecting the EAL of steel frames, whereas for RC frames, $A_{c,avg}$, and $\rho_{b,avg}$ show higher impact on predictions. The high importance of these parameters aligns with design principles, as steel frame sections for beams and columns are selected based on their section modulus ($Z$), which is subsequently related to the moment of inertia. Similarly, for RC frames, the cross-sectional area and longitudinal reinforcement are among the primary design parameters. The higher importance of floor area and height-related parameters for both inventories is also intuitive, as these parameters determine the taxonomy that is used to generate consequence and fragility thresholds in the HAZUS method.

Figure 3 shows the ALE plots of key features for each inventory. These parameters were selected due to (i) Shapely plots (Figure 2.(c) and 2.(d)) suggesting their higher importance, (ii) and their relevance to the conventional design of framed structures (e.g., moment of inertia can be used to change member sizing). For steel buildings, the EAL increases with $NS$ up to 14-story buildings, and then drops for the tallest frames (i.e., 19-story buildings). While it is intuitive that loss increases for taller buildings, the decrease in loss between 14- and 19-story buildings can be explained by the competition between the effect of height on seismic forces and the number of damageable components. For RC inventory, the increase in loss with $NS$ is also observed, although the slope is gentler due to the narrower range of story numbers (between 3-6 stories, compared to 1-19 for the steel frame). Both inventories show a linear increase in loss with $A_F$, as loss is calculated by multiplying normalized loss by the square unit price and gross floor area.

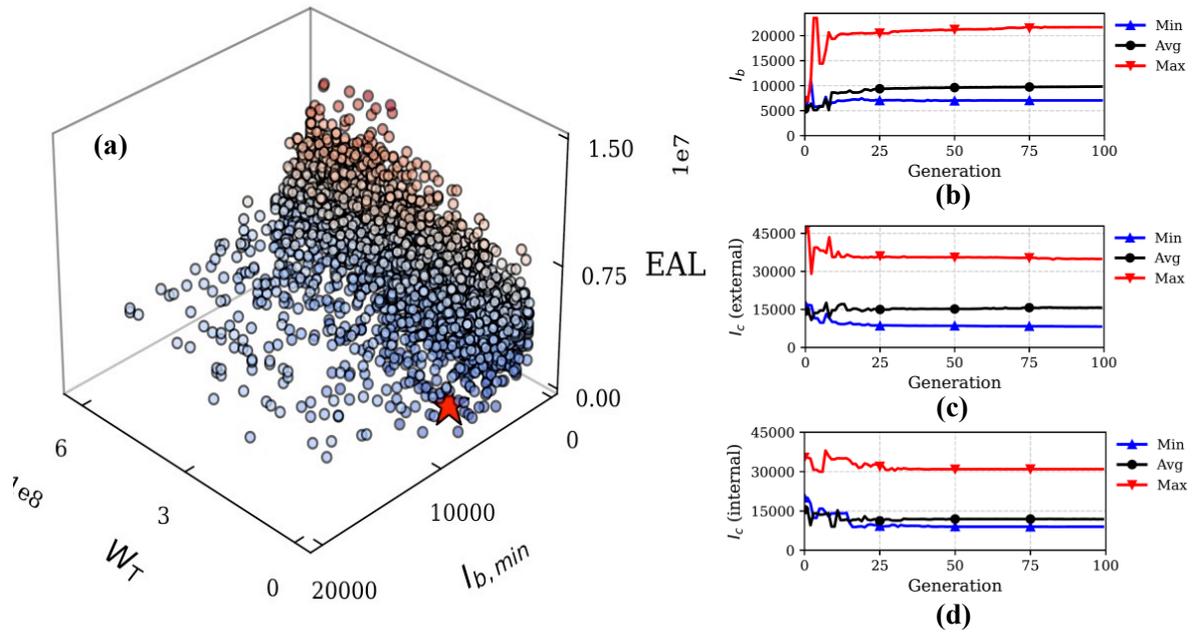

**Figure 4.** Genetic optimization using trained SVM for steel frame database: (a) populating the design space for a 5-story building with fixed floor area, optimizing moment of inertia of (b) beams, (c) external columns, and (d) internal columns over 100 generations.

For design-related parameters of steel frame inventory, the increase in internal columns' moment of inertia-related parameters (maximum and average; $I_{c,I,max}$ and $I_{c,I,avg}$, respectively) increases loss values, where this increase flattens for large values of these parameters. On the other hand, the increase of $I_{b,min}$ reduces the loss for $I_{bmin} < 10^4 \ in^4$, and increases the loss at $I_{b,min} > 10^4 \ in^4$. This observation suggests that while using larger sections for steel columns can reduce EAL up to a certain threshold, there is no practical benefit beyond this threshold. Similarly, the designer should use beam sections larger than a certain minimum; however, using larger minimum values would actually increase EAL. Both of these observations conform to the weak-beam strong column concept for frame seismic design, where the designer would intentionally uses stronger columns (i.e., larger moment of inertia) and weaker beams (reduced moment of inertia) to place plastic hinges in the beams for a ductile yielding behavior. For RC frames, the increase of $A_{b,avg}$ and $A_{c,avg}$ reduces loss, where the effect is more pronounced for $A_{b,avg}$. The increase in beams' longitudinal reinforcement results in lower loss, where the effect diminishes for $\rho_{b,avg} > 1.5\%$, similar to recommended reinforcement values for beam design.

Figure 4 illustrates the application of the trained SVM models to solve the inverse problem of determining the member sectional information for minimizing the EAL of a 5-story steel building. The building has 5 bays of 30 $ft$ in each direction

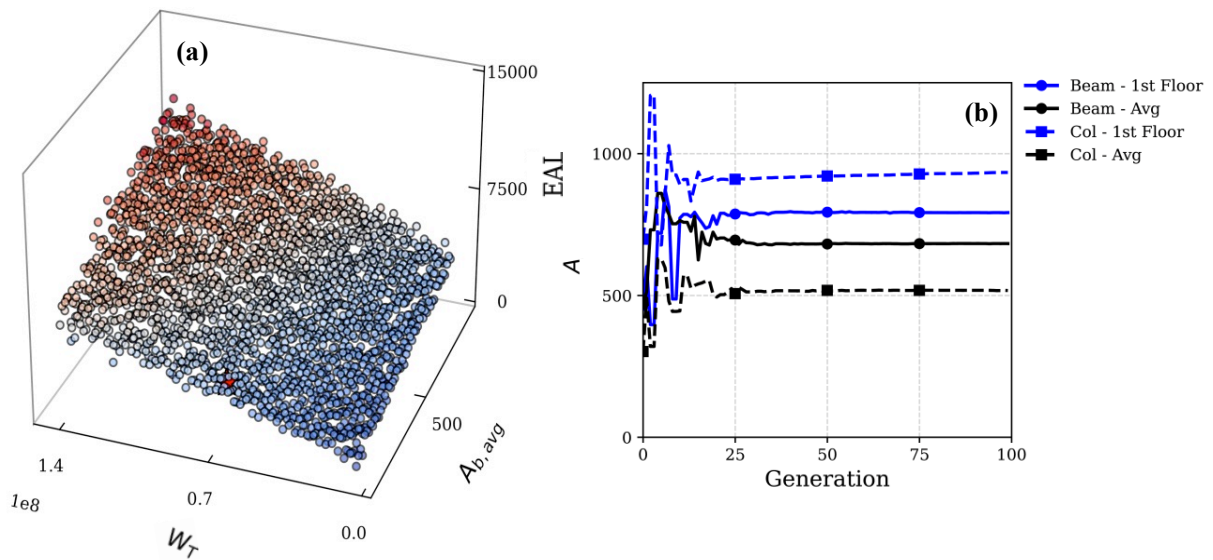

**Figure 5.** Genetic optimization using trained SVM for RC frame database: (a) populating the design space for a 5-story building with fixed floor area, (b) optimizing beams and columns cross-sectional area

(a total floor area of 22500 $ft^2$) and a first-story height of 13 $ft$. Figure 4.(a) shows the design space populated by the trained SVM over different ranges of $W_T$ and $I_{b,min}$, where other design parameters were kept at the average value of the database. As shown in Figures 4.(b)-4.(d), the optimization process converged within fewer than 30 generations, identifying an optimal design (i.e., lowest EAL for this geometry and structure type), characterized by the lower average moment of inertia for beams (approximately 66.7% of columns) that conform to the principle of strong-column weak-beam of moment frames. In addition, for both beams and columns, the average values of the moment of inertia are closer to the minimum values than the maximum values, where the difference is larger for external columns. This observation suggests that the optimized design was achieved by minimizing section sizes (to reduce seismic weight), where external columns were selected to be stronger than internal columns to provide enhanced lateral stability.

Figure 5 illustrates the optimization process for a 5-story reinforced concrete (RC) building with the same fixed geometry. Similarly, the optimization process converged within 30 generations. Figure 5.(a) demonstrates the design space based on $W_T$ and $A_{b,avg}$. Figure 5.(b) illustrates that the beam and columns' cross-sectional area for the first floor is larger than the average values over the entire building, where the columns of the first floor have a larger area than the corresponding beams, which is significantly higher than the average column area. The use of stronger sections in the first floor aligns with the vertical seismic force distribution of buildings governed by the first mode, where higher seismic forces should be resisted at lower stories.

## 5. Conclusion

### 5.1. Summary

This paper introduces a framework for integrating explainable machine learning models and optimization techniques to handle performance-based design as an inverse problem. The framework was applied to two different steel and RC moment frame databases in two geographical regions with different seismicity. For both databases, sectional information on frame members was obtained to minimize the expected annualized loss in terms of repair costs due to earthquake damage for a given geometry. The results showed that the trained support vector machines achieved an adjusted $R^2$ of more than 90% on the test for both building databases, using only geometry and design-related parameters. The number of stories and floor area were deemed influential geometric features, irrespective of building type. For steel frames, the moment of inertia of beams and columns has a notable impact on model prediction, whereas the cross-sectional area of beams and columns and beams' longitudinal reinforcement ratios impact the EAL of RC frames. All the dominant features conform to engineering principles of designing RC and steel moment frames.

Incorporating the trained SVM as the fitness function into the genetic algorithm facilitated the identification of optimized sectional information that minimized EAL within a few generations and at a low computational cost. Additionally, the inferred design information aligns with engineering principles, such as the strong-column weak-beam approach or the strategic placement of larger sections in areas subjected to higher seismic forces. These observations support the idea that data-driven models capture, at least partially, the essential physics that governs the seismic performance of framed structures. Furthermore, the accuracy and consistency of results across two different databases with diverse structural systems and designs suggest possible generalizability of the framework.

### 5.2. Limitations and future work

Despite the practical potential of the presented methodology, several limitations warrant further research in this area. First, while the presented methodology demonstrates the feasibility of creating surrogate models that generalize over a fixed taxonomy (a class of buildings with a fixed lateral system and moderate variations in their topology) under a specific hazard, such models cannot generalize beyond the trained samples. In addition, some recent work suggests that such models may be sensitive to small changes to the design space on which they were trained [22]. Such limitations motivate the exploration of possible remedies for generalizability, including the embedding of physical constraints [50,51], developing suites of surrogate models, transfer learning [52], or combining generative AI with surrogate modelling [53]. Another limitation of such models is the limited availability of databases that can support the development of data-driven surrogate models, where existing ones often provide limited topologies, design information, or require significant additional work. Developing such databases and providing clear schemas and documentation through public data repositories would be an important step in improving current methodologies. Future work is also needed to compare the output of such a framework to real design solutions and ensure their compliance with code requirements or optimality. In addition, currently, no benchmark exists to illustrate the level of model accuracy required to ensure that the final design solutions provide feasible and appropriate solutions. Such issues motivate larger collaboration between practicing engineers and academia to develop benchmarking processes. Lastly, the framework relies on the methodological choices used to derive the simulation data, including loss assessment methodologies. While previous work supports the feasibility of developing ML models irrespective of these choices, the accuracy and precision of target performance metrics depend on such assumptions.

## Data Availability Statement

All data and models of this study will be published in the following repository: https://github.com/StoStruct/InverseProblem

## Acknowledgement

No funding was received for this study.